\documentclass[letterpaper, 10 pt, conference]{ieeeconf}  

\IEEEoverridecommandlockouts                              

\usepackage{graphics} 
\usepackage{graphicx}
\usepackage{booktabs}
\usepackage{pifont}
\usepackage{amsfonts}
\usepackage{amsmath}
\usepackage{times}
\usepackage{float}
\usepackage{multicol}
\usepackage{multirow}
\usepackage{array}
\usepackage{booktabs}
\usepackage[dvipsnames]{xcolor}
\usepackage{color, colortbl}
\usepackage{caption}

\newcommand{\Loss}{\mathcal{L}}
\newcommand{\cmark}{\ding{51}}%
\newcommand{\xmark}{\ding{55}}%

\DeclareMathOperator*{\softmax}{softmax}
\DeclareMathOperator*{\argmax}{argmax}
\DeclareMathOperator*{\mean}{mean}

\title{\LARGE \bf
Multi-modal NeRF Self-Supervision for LiDAR Semantic Segmentation
}

\author{Xavier Timoneda$^{1}$, Markus Herb$^{1}$, Fabian Duerr$^{1}$, Daniel Goehring$^{2}$, Fisher Yu$^{3}$ 
\thanks{$^{1}$Xavier Timoneda, Markus Herb, and Fabian Duerr are with the Onboard Fusion team at CARIAD SE,
        Volkswagen Group, Ingolstadt, Germany.
        {\tt\small xavier.timoneda.comas@cariad.technology}}%
\thanks{$^{2}$Daniel Goehring, is with the Dahlem Center for Machine Learning and Robotics group at Freie Universit{\"a}t Berlin, Germany.
        }%
\thanks{$^{3}$Fisher Yu is with the Department of Information Technology and Electrical Engineering at ETH Z{\"u}rich, Switzerland
        }%
}

\begin{document}

\maketitle

\thispagestyle{empty}
\pagestyle{empty}

\begin{abstract}
LiDAR Semantic Segmentation is a fundamental task in autonomous driving perception consisting of associating each LiDAR point to a semantic label.
Fully-supervised models have widely tackled this task, but they require labels for each scan, which either limits their domain or requires impractical amounts of expensive annotations. 

Camera images, which are generally recorded alongside LiDAR pointclouds, can be processed by the widely available 2D foundation models, which are generic and dataset-agnostic.
However, distilling knowledge from 2D data to improve LiDAR perception raises domain adaptation challenges. For example, the classical perspective projection suffers from the parallax effect produced by the position shift between both sensors at their respective capture times. 


We propose a Semi-Supervised Learning setup to leverage unlabeled LiDAR pointclouds alongside distilled knowledge from the camera images. To self-supervise our model on the unlabeled scans, we add an auxiliary NeRF head and cast rays from the camera viewpoint over the unlabeled voxel features.
The NeRF head predicts densities and semantic logits at each sampled ray location which are used for rendering pixel semantics. Concurrently, we query the Segment-Anything (SAM) foundation model with the camera image to generate a set of unlabeled generic masks. We fuse the masks with the rendered pixel semantics from LiDAR to produce pseudo-labels that supervise the pixel predictions. During inference, we drop the NeRF head and run our model with only LiDAR. 

We show the effectiveness of our approach in three public LiDAR Semantic Segmentation benchmarks: nuScenes, SemanticKITTI and ScribbleKITTI.





\end{abstract}

\begin{figure}[h]
    \centering
    \includegraphics[width=0.49\textwidth]{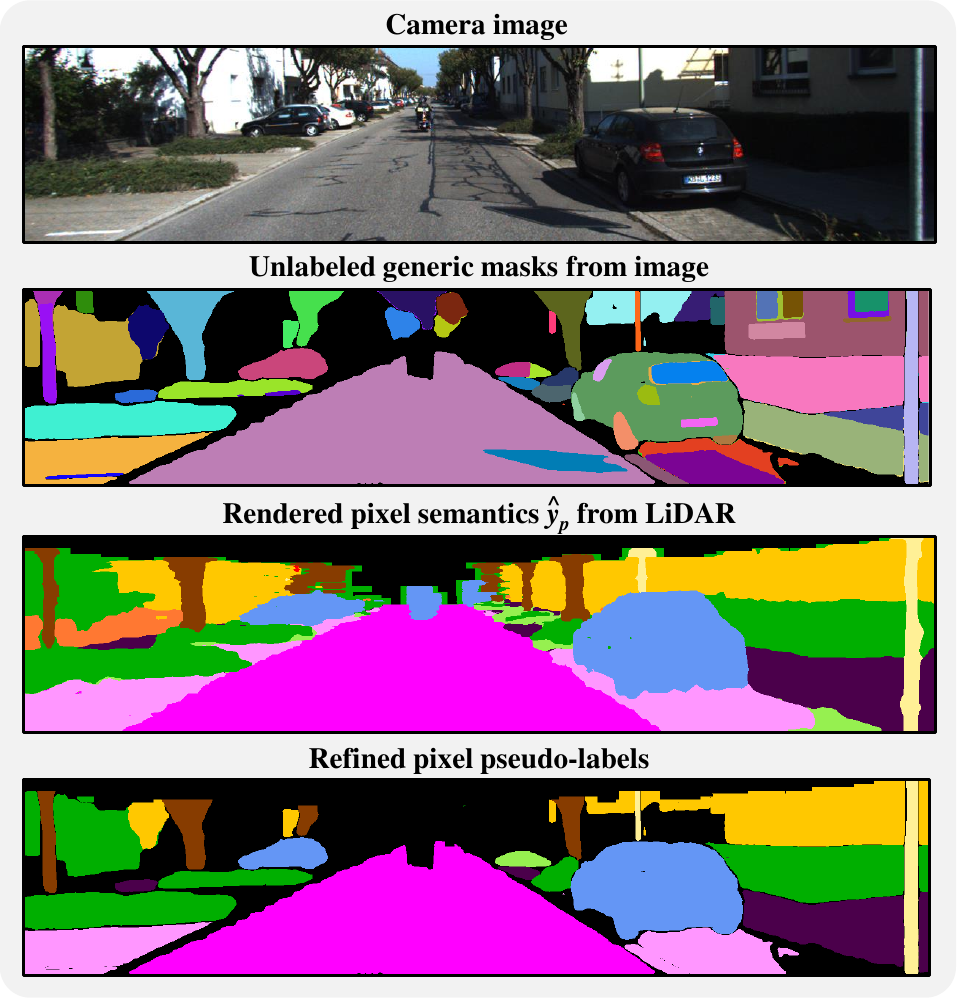}
    \caption{\textbf{Pixel pseudo-label generation.} Our Semi-Supervised setup leverages unlabeled multi-modal data by casting rays from the camera viewpoint into the unlabeled LiDAR features to render pixel semantic predictions. The renderings are fused with unlabeled generic masks from SAM foundation model to produce confident, refined pixel pseudo-labels.}
    \label{fig:initialfigure}
\end{figure}

\section{INTRODUCTION}

LiDAR Semantic Segmentation (SemSeg) is the computer vision task of associating every sample from a LiDAR pointcloud to a semantic class. It is essential for autonomous driving, enabling the identification of drivable areas and object boundaries to make safe and informed decisions when navigating the vehicle's surroundings \cite{Roriz2022} \cite{Pendleton2017}. 


The availability of autonomous driving perception benchmarks with annotated LiDAR scans \cite{semantickitti} \cite{scribblekitti} \cite{nuscenes} \cite{waymoopen} has enabled fully-supervised LiDAR SemSeg models \cite{SalsaNext} \cite{PolarNet} \cite{AMVNet} \cite{Tang2020} to learn this task for narrow domains (e.g. specific dataset, selected cities/countries) where only limited amounts of labeled data are enough to train a well-performing perception model. However, for bringing fully automated driving functions to a world scale, we need models that perform well in a wide variety of scenarios, such as distinct continents, road agents, weather conditions, or driving styles. In the fully-supervised domain, this involves labeling huge amounts of LiDAR pointclouds, which is an expensive task that requires human supervision \cite{Mahony2019}. 

On the other hand, LiDAR pointclouds and camera images are cheap to collect compared with their annotation cost \cite{Schon2009}. Generally, and for most of the public LiDAR perception benchmarks, LiDAR pointclouds are recorded alongside synchronized camera images at similar instants, with significant overlap in their field of view and a known transformation between the two sensor origins \cite{semantickitti} \cite{scribblekitti} \cite{nuscenes} \cite{waymoopen}.

The camera frames, unlike LiDAR pointclouds, can be processed by the widely available 2D foundation models \cite{dinov1} \cite{dinov2} \cite{sam}, which are increasingly effective at helping with a wide range of tasks while being dataset agnostic. 
The Segment-Anything Model \cite{sam} (SAM) is a groundbreaking foundation model designed for zero-shot image segmentation. For each image, it generates masks that are agnostic to the instance definition and labeling format. 
Our work aims to distill the knowledge from SAM for the LiDAR modality during training, while having a LiDAR-only inference pipeline. Fig. \ref{fig:initialfigure} exemplifies this knowledge distillation.

However, distilling knowledge from 2D foundation models into a 3D perception model raises the challenge of effectively bridging these data from distinct domains. There's been extensive work tackling this challenge. The classical way of fusing 2D and 3D data, already known for over a decade, is the perspective projection \cite{Hughes2013} of LiDAR points into the camera plane given the camera parameters and transformation between sensors. This technique has been demonstrated useful in the past for assigning 3D pseudo-labels from 2D semantic labels, which are typically less expensive to produce, requiring less skilled annotators and more generic equipment \cite{Mahony2019}. 

Even though the perspective projection has proven effective and accurate in some scenarios, it suffers from the parallax effect, which is the apparent shift in the position of an object relative to a background when the observer's viewpoint changes \cite{Ginsburg2008}. In our multi-modal setup, this occurs when the position of the LiDAR and camera at their respective capture times differ. This undesired effect becomes especially stronger at higher velocities and in setups where the camera is placed away from the LiDAR.

Some works address this issue by exploiting the multi-view nature of video recordings at consecutive frames or by overlapping multiple-view cameras to select projections that are consistent from multiple views \cite{Genova2021}. However, these works require pre-trained 2D SemSeg models and need to either train on data with multi-view cameras and big overlaps or deal with moving objects from neighboring frames.

To address this, we include in our Semi-Supervised Learning (SSL) model a self-supervision technique to train on the unlabeled data which is inspired by the training mechanism of Neural Radiance Fields (NeRFs) \cite{nerf}. We show how this technique can better leverage unlabeled images and scans than existing approaches such as the classical perspective projection. Our NeRF self-supervision allows the model to reason about occupancy and semantics along rays, rather than individual points. It consists of 1) an efficient Pixel-to-Ray Casting mechanism, 2) a $NeRF$ Multi-Layer Perceptron (MLP) head, and 3) volumetric rendering equations. The whole NeRF self-supervision is only added during training and dropped for inference. Hence, once deployed, our model is as efficient as any LiDAR-only SemSeg method and doesn't require any input data other than LiDAR. We demonstrate the benefit of our method in scenarios with scarcely labeled 3D data on three well-established public benchmarks for LiDAR SemSeg: nuScenes \cite{nuscenes}, SemanticKITTI \cite{semantickitti}, and its scarcely-labeled variation: ScribbleKITTI \cite{scribblekitti}.

\section{RELATED WORK}

\textbf{LiDAR Semantic segmentation.} It has been tackled with different mechanisms. Range view methods \cite{SalsaNext} \cite{Milioto2019} \cite{Xu2021} \cite{Zhao2021} project the points into a range image and process it with convolutional neural networks equivalent to 2D SemSeg networks \cite{Chen2017}. Even though their lightweight advantage, the pointcloud is not processed in its original 3D domain, causing undesired effects such as semantic incoherence or shape deformation \cite{Kong2023}. Point-based methods, which operate directly on the raw LiDAR pointclouds, gained some popularity with the advent of PointNet \cite{PointNet}. Recently, bird’s eye view \cite{Zhang2020} and multi-view \cite{AMVNet} \cite{RPVNet} methods have also been proposed. Alternatively, voxel-based methods \cite{Zhao2020} \cite{Cylinder3D2020} \cite{Cylinder3D2021} discretize the continuous volume covered by the LiDAR into voxels, which capture the information contained in its points. This enables efficient sparse operations while processing the poincloud in its original domain. We adopt Cylinder3D \cite{Cylinder3D2021}, a voxel-based method widely used in the literature, as the baseline architecture to run our experiments.

Extensive experiments on all the aforementioned methods show very appealing results for the fully-supervised setup on several LiDAR SemSeg benchmarks, but their performance significantly degrades in the low-data regime \cite{Gao2020}. Some recent works alleviate the label reliance with weak \cite{SQN} \cite{Zhang2021}, scribble \cite{scribblekitti} and box \cite{Box2Seg} supervision strategies that reduce the annotation cost. We propose a semi-supervised learning setup that can leverage large amounts of easy-to-acquire unlabeled data to boost the model's performance.

\definecolor{Gold}{rgb}{1.0,0.75,0.0}
\begin{figure*}[t]
\centering
\includegraphics[width=0.99\textwidth]{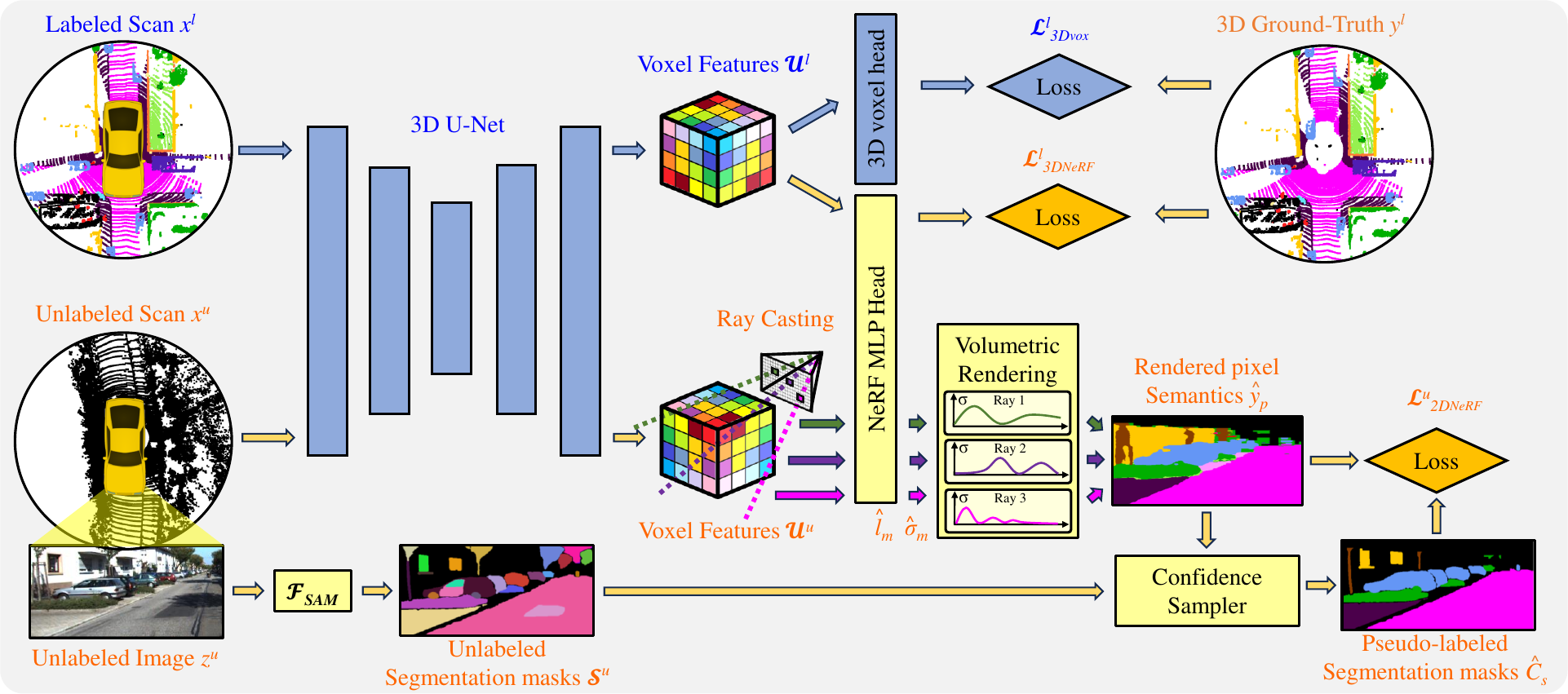}
\caption{\textbf{Overview of our method.} During training, labeled $x^{l}$ and unlabeled $x^{u}$ scans are processed in parallel by a 3D U-Net. The voxel features $\mathcal{U}^{l}$ of the labeled scan are processed by both $vox_{3D}$ and $NeRF$ heads to get point-wise semantic predictions. These are compared against the 3D Ground-Truth $y^{l}$ to form the supervised 3D losses $\Loss_{3D_{vox}}^{l}$ and $\Loss_{3D_{NeRF}}^{l}$. Concurrently, for each unlabeled image $z^{u}$ from $x^{u}$, we obtain generic masks $\mathcal{S}^{u}$ with ${F}_{SAM}$ foundation model. We trace $P$ rays from the camera origin with each pixel's direction, and sample the unlabeled voxel features $\mathcal{U}^{u}$ at $M$ locations along each ray. These are processed by the $NeRF$ head to predict $P \times M$ semantic logits $\hat{l}_{m}$ and densities $\hat{\sigma}_{m}$. We integrate $\hat{l}_{m}$ and $\hat{\sigma}_{m}$ along the ray to render the per-pixel class probabilities $\hat{y}_p$. The confidence sampler merges $\hat{y}_p$ with the masks $\mathcal{S}^{u}$ to get refined pseudo-labels $\hat{\mathcal{C}}_{s}$. The predictions $\hat{y}_p$ are compared against the pseudo-labels $\hat{\mathcal{C}}_{s}$ to form the self-supervised 2D loss $\Loss_{2D_{NeRF}}^{u}$. At inference time, we remove all components represented in \textcolor{Gold}{\textbf{yellow}}, resulting in a LiDAR-only inference.}
\label{fig:overallpipeline}
\end{figure*} 

\textbf{Semi-Supervised Learning.} SSL methods operate in partially labeled datasets, where large amounts of unlabeled data are combined with small amounts of labeled data to boost the model's performance.
In image SemSeg, some works leverage unlabeled data by enforcing consistency constraints between predictions \cite{Chen2021} \cite{Ouali2020} \cite{MeanTeacher} or weights \cite{Ke2020} from two perturbed views of the same image. 
Self-learning \cite{Lee2013} is a branch of semi-supervised learning in which a labeled subset of the data is used to train a teacher model that automatically generates pseudo-labels to train on the unlabeled data. Several image SemSeg works adopt this framework by generating pixel pseudo-labels on the unlabeled data to further train the network \cite{Hu2021} \cite{Ke2020} \cite{Yang2022} \cite{Yuan2021}. LaserMix \cite{LaserMix} is one of the first works to explore SSL for LiDAR SemSeg. Their self-learning system encourages the model to make consistent and confident predictions from unlabeled scans by exploiting a spatial prior.

\textbf{Knowledge Distillation.} In Machine Learning, this term denotes the process of compressing the knowledge present in large, complex models into smaller task-specific models while holding their performance. The principle of distillation involves training a 'student' model to imitate the original or 'teacher' model \cite{Bang2021}. In many cases, the performance of the student exceeds the teacher's in the specific task. Foundation models are very suitable for being used as teachers in knowledge distillation, as they are trained on huge amounts of diverse data and are valid for a wide range of tasks \cite{Zhou2023}. In the field of computer vision, foundation models with Vision Transformer (ViT) architectures have revolutionized the traditional computer vision tasks \cite{dinov1} \cite{dinov2}. The recently published Segment-Anything Model (SAM) \cite{sam} is a novel image segmentation model, trained with over 1 billion masks on 11M images. It's designed for zero-shot image segmentation, producing masks for instances present in an image, while being agnostic to the instance definition and annotation format. 

\textbf{Multi-modal distillation.} Within this work's context, we refer to this term as the process of transferring the knowledge from a model that uses both camera and LiDAR unlabeled data into an unimodal LiDAR SemSeg model that is used for inference. The challenge of bridging the gap between both domains has been tackled in the past \cite{Huang2022}. The classic way of fusing 3D and 2D data is the perspective projection \cite{Hughes2013} of LiDAR points into the camera plane with the intrinsics and transformation between sensors. The projected points are assigned to the corresponding image labels. This technique has been demonstrated useful for assigning 3D pseudo-labels from labeled images. The work in \cite{Genova2021} generates image pseudo-labels of a sequential dataset with a pre-trained image SemSeg model and obtains 3D pseudo-labels with perspective projection. It alleviates the undesired parallax effect by enforcing multi-view consistency from distinct viewpoints. Unlike our approach, this work requires the input data to be sequential, as well as a pre-trained image SemSeg model trained with the same dataset and labeling format to produce the 2D pseudo-labels. In the reversed direction, the work \cite{Xie2016} draws 3D bounding primitive annotations and automatically pseudo-labels images by enforcing prediction coherence between sequential frames.

\section{METHOD}

In this section, we establish our problem mathematically and provide a detailed description of every component of our Semi-Supervised LiDAR Semantic Segmentation system.

\textbf{Problem formulation.} Given a partially labeled Dataset $\mathcal{D} = \left\{\mathcal{D}^{l},\mathcal{D}^{u}\right\}$, containing $N^{l}$ labeled scans $x^{l}$ with unlabeled images $z_{n}$: $\mathcal{D}^{l} = \left\{\left(x_{n}^{l}, y_{n}^{l}\right), z_{n}\right\}_{n=1}^{N^{l}}$ and $N^{u}$ unlabeled scans $x^{u}$ with unlabeled images: $\mathcal{D}^{u} = \left\{x_{n}^{u}, z_{n}\right\}_{n=1}^{N^{u}}$, we aim to leverage both labeled and unlabeled sets to train a LiDAR SemSeg network. Concurrently, we make use of the Segment-Anything foundation model $\mathcal{F}_{SAM}$ that provides $K$ generic masks $\mathcal{S}^{u}$ for every unlabeled image.

\textbf{System overview.} Fig. \ref{fig:overallpipeline} shows the pipeline of our SSL system during training. We adopt Cylinder3D \cite{Cylinder3D2021} as the baseline architecture for our experiments. We follow the typical network structure for voxel-based 3D LiDAR SemSeg networks, with a sparse 3D U-Net backbone followed by a voxel 3D SemSeg head $vox_{3D}$ with a convolutional 3D layer. The rest of the model components are only present during training and removed for inference.

\begin{figure}[h]
    \centering
    \includegraphics[width=0.49\textwidth]{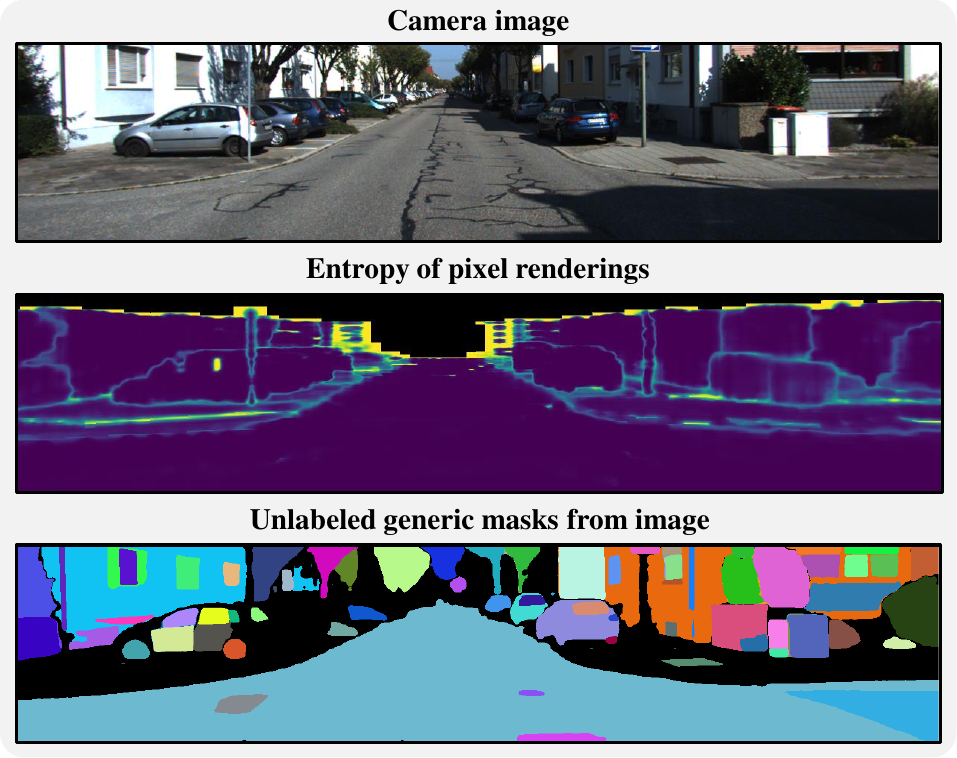}
    \caption{\textbf{Entropy of the pixel renderings.} In this context, we use the entropy $\mathcal{H}$ as a measure of the uncertainty in the predicted class probabilities of each rendered pixel. Bright colors denote high entropy. We observe the highest entropy $\mathcal{H}$ concentrates at object boundaries.}
    \label{fig:entropy}
\end{figure}

During training, every input batch contains both labeled $x^{l}$ and unlabeled $x^{u}$ scans. These are fed to the U-Net to obtain the voxel features $\mathcal{U}^{l}$, $\mathcal{U}^{u}$ from both labeled and unlabeled scans. The voxel features of the labeled scan $\mathcal{U}^{l}$ are fed to both $vox_{3D}$ and $NeRF$ heads. The latter consists of a Multi-Layer Perceptron that predicts semantic logits $\hat{l}_{m}$ and density $\hat{\sigma}_{m}$ for every sampled location $m \in \mathbb{R}^{3}$. For the labeled cue, every sampled location $m$ is the coordinates of each point in the labeled scan. The semantic predictions of both $vox_{3D}$ and $NeRF$ heads are compared against the 3D ground truth labels $y^{l}$, producing the 3D supervised semantic losses $\Loss_{3D_{vox}}^{l}$ and $\Loss_{3D_{NeRF}}^{l}$. These losses are used to train the $vox_{3D}$ and $NeRF$ heads with the labeled subset.

\textbf{Ray Casting.} Concurrently, we sample $P$ pixels from the unlabeled image $z^{u}$ and trace a ray $r_{p}$ for each of them. For every sampled pixel $p$, we obtain its camera origin and pixel's direction in the LiDAR coordinates by using the known transformation and camera parameters. With the per-pixel origin and direction, we cast $r_{p}$ and sample the unlabeled voxelized features $\mathcal{U}^{u}$ at $M$ uniformly sampled locations along $r_{p}$ between the $\Pi_{near}$ and $\Pi_{far}$ planes. We use trilinear interpolation to obtain the sampled feature from $\mathcal{U}^{u}$ at each location. Adding explicit positional encodings of the sampled locations didn't improve the model's performance, which suggests the network is implicitly using the interpolated positional embeddings of the voxels.

To improve the pixel sampling efficiency, we only sample the minimum amount of pixels ensuring that each visible voxel by the camera is traversed by at least one ray. This allows bigger batch sizes thus speeding up the whole training pipeline with negligible performance degradation.

The $NeRF$ head predicts semantic logits $\hat{l}_{m}$ and density $\hat{\sigma}_{m}$ for each sampled location $m$. We then obtain per-pixel semantic logits $\hat{l}_{p}$ from $\hat{l}_{m}$ and $\hat{\sigma}_{m}$ with volumetric rendering.

\textbf{Volumetric Rendering.} We follow the standard structure used for pixel color rendering when querying a NeRF for a novel view of the scene. First, the opacity $\alpha$ of each sampled location $m$ along the ray is estimated from the density $\hat{\sigma}_{m}$ and the separation between consecutive locations $\delta_{m}$:

\begin{equation}
\hat{\alpha}_{m} = 1 -exp(-\hat{\sigma}_{m}\delta_{m}) \qquad \hat{T}_{m}=\prod_{j=1}^{m-1} (1-\hat{\alpha}_{j})
\label{eq:alphatransmittance}
\end{equation}

while the transmittance $\hat{T}_{m}$ is computed as the cumulative product of the estimated opacities along the ray. We render the semantic logits $\hat{l}_{p}$ of each pixel and obtain the class probabilities $\hat{y}_{p}$ with $\softmax$ normalization:

\begin{equation}
\hat{l}_{p}=\sum_{m=1}^{M} \hat{T}_{m}\hat{\alpha}_{m}\hat{l}_{m} \qquad \hat{y}_{p} = \softmax_c(\hat{l}_{p})
\label{eq:volrendering}
\end{equation}

 The NeRF self-supervision, including ray casting, $NeRF$ MLP head, and volumetric rendering is a crucial component of our system, as it enables effectively bridging the 3D LiDAR data and the 2D distilled knowledge from images.

\textbf{Confidence Sampler.} Concurrently, we query the pre-trained ViT-H Segment-Anything Model (SAM) \cite{sam} to obtain the unlabeled masks $\mathcal{S}^{u}$ for every segment of the camera image from the unlabeled scan. We feed them into the confidence sampler along with the rendered pixel semantics $\hat{y}_{p}$ to produce the segment-wise 2D pseudo-labels $\hat{\mathcal{C}}_{s}$. 

Fig. \ref{fig:entropy} shows how the entropy $\mathcal{H}$ (uncertainty) of the rendered pixel semantics $\hat{y}_{p}$ concentrates at the object boundaries. Our confidence sampler benefits from this effect by leveraging confident pixel predictions at the interiors of each segment to supervise the unreliable predictions at the borders. 

For producing $\hat{\mathcal{C}}_{s}$, the confidence sampler takes all rays belonging to $S$ and computes the $\argmax$ of the probabilities $\hat{Y}_{p} = \argmax(\hat{y}_{p})$ over the $C$ classes. The segment pseudo-label $\hat{\mathcal{C}}_{s}$ is computed as the $\argmax$ over the predictions of all pixels belonging to this segment $\hat{\mathcal{C}}_{s} = \argmax(\hat{Y}_{p})$.

We only keep the confident segment pseudo-labels $\hat{\mathcal{C}}_{s}$. We measure the pseudo-label confidence $\mathcal{H}_{\mathcal{S}}$ as the entropy of the resulting distribution when averaging all the probabilities across every ray belonging to the segment whose predicted label $\hat{Y}_{p}$ coincides with the segment pseudo-label $\hat{\mathcal{C}}_{s}$.

\begin{equation}
\mathcal{H}_{\mathcal{S}} = -\sum_{c=1}^{C} \bar{y}_{s} \log(\bar{y}_{s}), \quad \bar{y}_{s} = \mean_p(\hat{y}_{p}) \mid \hat{Y}_{p} = \hat{\mathcal{C}}_{s}
\label{eq:entropy}
\end{equation}

We only keep segment pseudo-labels $\hat{\mathcal{C}}_{s}$ with entropies lower than a threshold $\mathcal{H}_{th}$.

\begin{figure*}[t]
\centering
\includegraphics[width=0.99\textwidth]{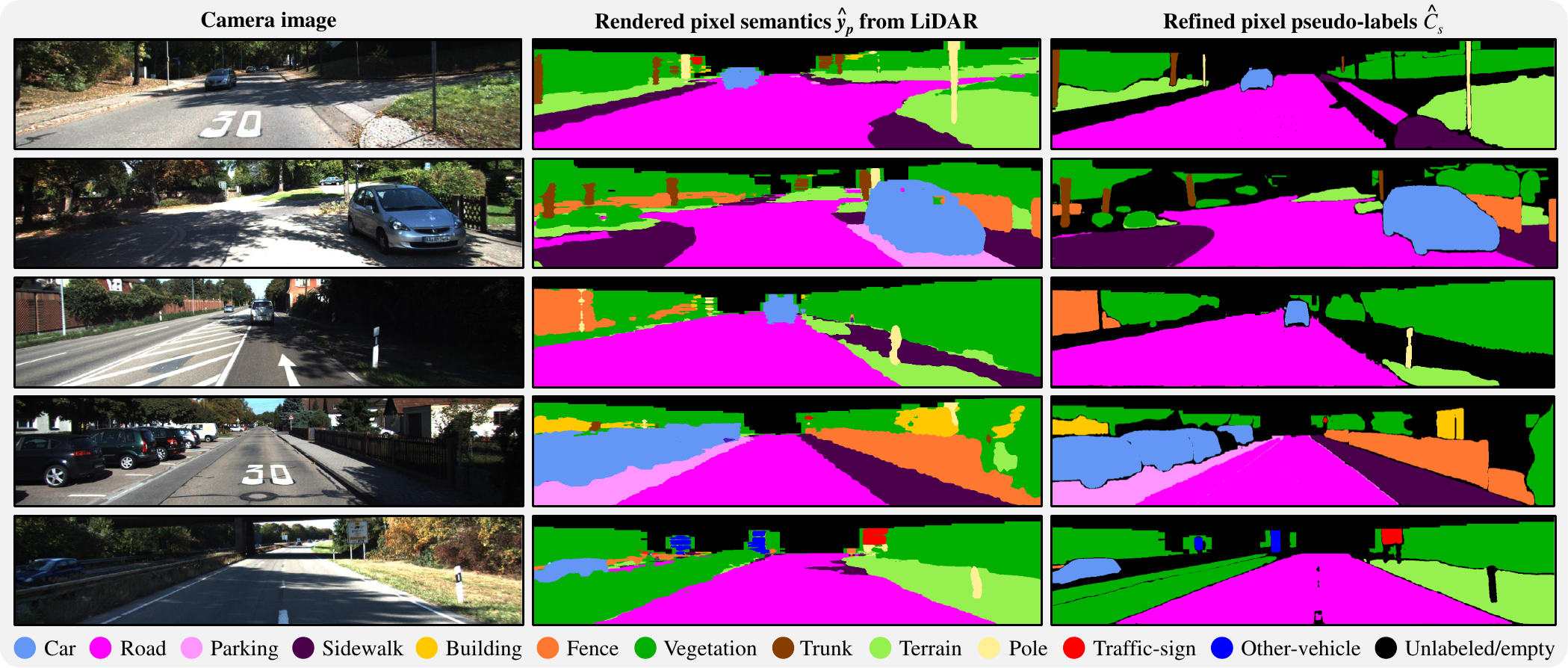}
\caption{\textbf{Qualitative analysis.} For each example of SemanticKITTI \cite{semantickitti}, we show the rendered pixel semantics $\hat{y}_{p}$ from the unlabeled voxel features and the generated pseudo-labels $\hat{\mathcal{C}}_{s}$ that supervise the pixel predictions.}
\label{fig:qualitative2d}
\end{figure*}

\textbf{Loss Function.} The rendered pixel semantics $\hat{y}_{p}$ are compared pixel-wise against the pseudo-labels $\hat{\mathcal{C}}_{s}$ to produce the 2D self-supervised semantic loss $\Loss_{2D_{NeRF}}^{u}$. The total loss $\Loss$ is a combination of the supervised 3D voxel and NeRF losses, and the self-supervised 2D loss:

\begin{equation}
\Loss = \beta\Loss_{3D_{vox}}^{l} + \gamma\Loss_{3D_{NeRF}}^{l} + \lambda\Loss_{2D_{NeRF}}^{u}
\label{eq:loss}
\end{equation}

where $\beta, \gamma, \lambda$ are weighting scalars. Each loss term is a weighted combination of the Cross-Entropy \cite{FCN} and the Lovasz \cite{Lovasz} losses:

\begin{equation}
\begin{aligned}
    \Loss_{x} = \mu\Loss_{CE}(y,\hat{y})  + \nu\Loss_{Lovasz}(y,\hat{y}) \\ 
    \Loss_{x} \in \left\{\Loss_{3D_{vox}}^{l}, \Loss_{3D_{NeRF}}^{l}, \Loss_{2D_{NeRF}}^{u}\right\}  \\ 
\end{aligned}
\label{eq:celovasz}
\end{equation}

where $y$ refers to the Ground-Truth annotation of each LiDAR point for $\Loss_{3D_{vox}}^{l}$ and $\Loss_{3D_{NeRF}}^{l}$ losses, and it refers to the pixel estimated pseudo-label for $\Loss_{2D_{NeRF}}^{u}$. The parameters $\mu$ and $\nu$ are also weighting scalars.

\textbf{Uni-Modal Inference:} At inference time, we keep only the baseline LiDAR-only architecture (3D U-Net and $vox_{3D}$ head) and remove everything else from the pipeline. The weights of the 3D U-Net, which were trained with both labeled and unlabeled data in a semi-supervised fashion, provide powerful and general voxel features that boost the model's performance.

\section{RESULTS}

\textbf{Data.} We evaluate our approach experimentally on the nuScenes \cite{nuscenes}, SemanticKITTI \cite{semantickitti} and ScribbleKITTI \cite{scribblekitti} benchmarks. nuScenes contains 28130 training scans and 6019 validation scans. Each scan has 6 camera images that jointly cover the whole LiDAR's angular range. We follow the official label format with 16 semantic classes. SemanticKITTI and ScribbleKITTI contain 19130 training scans and 4071 validation scans. Each scan has images from two frontal cameras operating as a single stereo camera. We only use the left frontal camera for the experiments on KITTI. We follow KITTI's official label format with 19 semantic classes. ScribbleKITTI is a variant of SemanticKITTI where only a subset of every training scan is annotated (approximately 8.06\%), leaving the rest of the points unlabeled. For all datasets, we adopt the same setup as in the literature on SSL LiDAR SemSeg \cite{LaserMix}. We randomly select 1\% and 10\% of the training scans for the labeled cue and assume the rest are unlabeled.

\definecolor{DarkBlueGrey}{rgb}{0.22,0.30,0.56}
\definecolor{Orange_colorblind}{rgb}{0.89,0.54,0.17}
\begin{figure*}[t]
\centering
\includegraphics[width=0.99\textwidth]{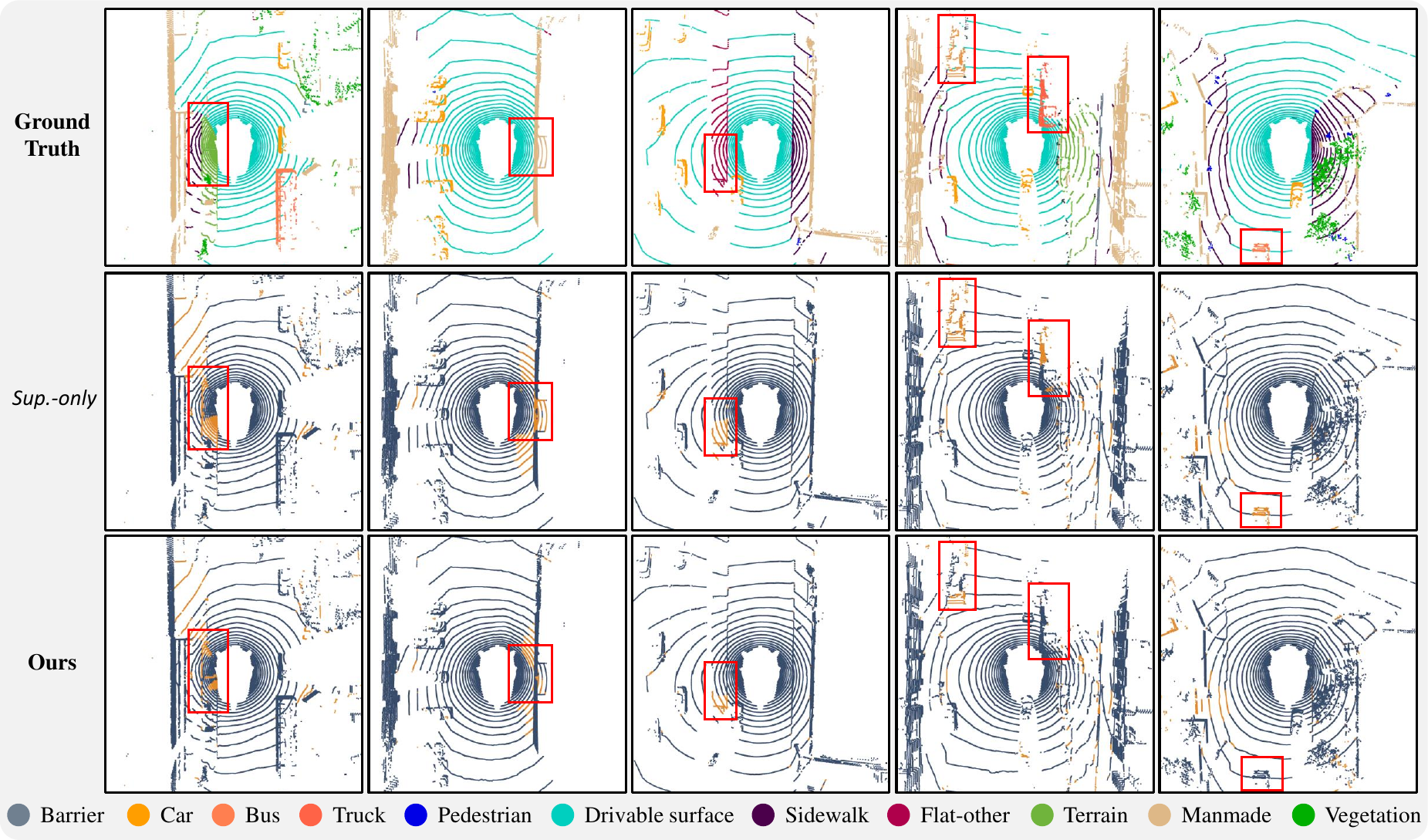}
\caption{\textbf{Qualitative results.} Error maps from LiDAR bird's eye view on 1\% split of nuScenes \cite{nuscenes}. The first row shows the \textbf{Ground-Truth} labels of each example. The second and third row show the \textcolor{DarkBlueGrey}{\textbf{correct}} and \textcolor{Orange_colorblind}{\textbf{incorrect}} predictions painted in \textcolor{DarkBlueGrey}{\textbf{blue}} and \textcolor{Orange_colorblind}{\textbf{orange}} for \textit{Sup.-Only} and \textbf{Ours} models, respectively. The \textcolor{Red}{\textbf{red}} boxes highlight the regions with notable differences.}
\label{fig:qualitative3d}
\end{figure*}

\textbf{Implementation Details.} We adopt Cylinder3D \cite{Cylinder3D2021} as the baseline architecture for our experiments. The voxel resolution is set to [240, 180, 20] for $radial$, $angular$ and $height$ coordinates, respectively. The weights of the U-Net and the semantic heads are randomly initialized. 
We set the number of sampled locations per ray $M$ to 458, and the $\Pi_{near}$ and $\Pi_{far}$ planes to 2.3 and 50 meters, respectively.
We use a 2-layer MLP with ReLU activation and 64 neurons for the first layer, and $C + 1$ outputs at the last layer: $C$ for the class logits and 1 for density. We apply a truncated exponential activation function to the density output.
We obtain the unlabeled masks $\mathcal{S}^{u}$ by querying the ViT-B HQ-SAM \cite{samhq} model with pre-trained weights from HQSeg-44K \cite{samhq}. HQ-SAM \cite{samhq} is a recent variation of SAM \cite{sam} with small architecture modifications and trained with 44K fine-grained image mask annotations which slightly improve the segmentation quality. To save computational effort, we pre-compute the segments for all the images of the dataset only once. We use SAM's automatic mask generation code provided by the creator's repository \cite{sam} to obtain $\mathcal{S}^{u}$ by randomly sampling $32\times32$ anchor points in the image. For the sake of keeping the masks as generic as possible, we leave the mask generator parameters to their default value.

We set the loss weighting factors $\beta$, $\lambda$, $\nu$ and $\mu$ constant to the values 0.5, 0.1, 1.0 and 3.0 respectively. For $\gamma$, we set a schedule that linearly decreases the value at every epoch from 1.0 to 0.
We set the entropy threshold $\mathcal{H}_{th}$ to 1.6 for nuScenes and 1.8 for SemanticKITTI and ScribbleKITTI. 

We denote the supervised-only baseline as \textit{sup.-only}. All experiments are implemented using PyTorch on NVIDIA Tesla V100 GPUs with 16GB of memory. 

\definecolor{LightCyan}{rgb}{0.95,0.95,1.0}
\definecolor{LightYellow}{rgb}{0.98,0.99,0.84}
\newcolumntype{C}[1]{>{\centering\arraybackslash}p{#1}}
\begin{table*}[!htbp]
    \centering
    \caption{Comparison of our approach with distinct Semi-Supervised LiDAR Semantic Segmentation models. The \textbf{best} and \underline{second best} score for each data split are highlighted with \textbf{bold} and \underline{underline}. Higher is better.}
    \begin{tabular}{C{2.1cm}|C{1cm}C{1cm}|C{1cm}C{1cm}|C{1cm}C{1cm}}
        \toprule
        \multirow{2}{*}{Method} & \multicolumn{2}{c|}{nuScenes \cite{nuscenes}} & \multicolumn{2}{c|}{SemanticKITTI \cite{semantickitti}} & \multicolumn{2}{c}{ScribbleKITTI \cite{scribblekitti}} \\
        & 1\% & 10\% & 1\% & 10\% & 1\% & 10\% \\
        \midrule
        \rowcolor{LightCyan}
        \textit{Sup.-only} & 50.9 & 65.9 & 45.4 & 56.1 & 39.2 & 48.0\\
        \midrule
        MeanTeacher \cite{MeanTeacher} & 51.6 & 66.0 & 45.4 & 57.1 & 41.0 & 50.1\\
        CBST \cite{CBST} & 53.0 & 66.5 & \underline{48.8} & 58.3 & 41.5 & 50.6\\
        CPS \cite{CPS} & 52.9 & 66.3 & 46.7 & \underline{58.7} & 41.4 & \underline{51.8}\\
        LaserMix \cite{LaserMix} & \underline{55.3} & \textbf{69.9} & \textbf{50.6} & \textbf{60.0} & \textbf{44.2} & \textbf{53.7}\\
        \midrule
        \textbf{Ours} & \textbf{55.9} & \underline{69.6} & 47.7 & 57.4 & \underline{43.3} & 51.6\\
        $\Delta$ $\uparrow$ & \textbf{\textcolor{Blue}{+5.0}} & \textbf{\textcolor{Blue}{+3.7}} & \textbf{\textcolor{Blue}{+2.3}} & \textbf{\textcolor{Blue}{+1.3}} & \textbf{\textcolor{Blue}{+4.1}} & \textbf{\textcolor{Blue}{+3.6}}\\
        \bottomrule
    \end{tabular}
    
\label{tab:sotacomparison}
\end{table*}

\textbf{Evaluation Strategy.} For every benchmark, we compare the performance of our method with the supervised baseline which only uses the labeled subset. We show how our SSL setup leads to performance upgrades with respect to the baseline. We also compare the results with Mean Teacher \cite{MeanTeacher}, CBST \cite{CBST}, CPS \cite{CPS} and LaserMix \cite{LaserMix}. For all the comparisons with the baseline and State-of-the-Art, we use the same labeled data splits and architecture. We adopt the standard mean Intersection over Union (mIoU) over all the classes of each benchmark as the metric to evaluate the segmentation performance of our method.

\textbf{Results on nuScenes.} Our SSL method improves the supervised-only baseline by 5.0\% and 3.7\% when using only 1\% and 10\% of the scan labels. Table \ref{tab:sotacomparison} shows how our model outperforms all the rest for the 1\% split and ranks second on the 10\% split, being very close to LaserMix \cite{LaserMix}, which ranks first.

\textbf{Results on SemanticKITTI.} Our method improves the supervised-only baseline by 2.3\% and 1.3\% when using 1\% and 10\% of the scan labels. Our model ranks third for both 1\% and 10\% splits, showing a significantly greater upgrade at the 1\% split.

\textbf{Results on ScribbleKITTI.} Our method improves the supervised-only baseline by 4.1\% and 3.6\% when using 1\% and 10\% of the scan labels. Our model ranks second for the 1\% split and ranks third for the 10\% split, being very close to CPS \cite{CPS}, which ranks second.

\textbf{Discussion.} The greatest performance upgrades are obtained in nuScenes. This matches our expectations since it's the only dataset in which the whole LiDAR's angular range is visible by at least one of the 6 multi-view cameras. For SemanticKITTI and ScribbleKITTI, we obtain smaller performance upgrades compared to other methods, mainly because the dataset has only one frontal camera that covers 22.5\% of LiDAR's angular range. For these benchmarks, only up to 22.5\% of each unlabeled scan can be leveraged by our SSL model. However, our method still shows competitive performance upgrades compared to the rest, which all use the full angular range of the unlabeled scans. 

The results also suggest our model delivers the highest improvements with lower amounts of labeled data. An explanation for this behavior is the existing projection errors between 3D and 2D coming from imperfect calibration. These errors might be more critical when trying to upgrade the performance of already well-performing models trained with bigger amounts of labeled data. In such scenarios, the bigger classes already perform well, and therefore the mIoU can only get significantly upgraded by improving the $IoU$ on classes with thin objects such as poles, signs, or pedestrians, which are the most affected by the imperfect calibration.

\textbf{Qualitative analysis.} Fig. \ref{fig:qualitative2d} shows a representation of the rendered pixel semantics and the refined pseudo-labels for several examples from the unlabeled training split in SemanticKITTI. It exemplifies our model's ability to automatically produce high-quality refined pseudo-labels from the rendered pixel semantics of unlabeled scans and the generic SAM masks from the camera image. We present additional qualitative results in our supplementary video.

Fig. \ref{fig:qualitative3d} shows a visual representation of our method's 3D semantic segmentation performance in several scans from the \textit{val} set of nuScenes \cite{nuscenes}. For each example, we represent the Ground Truth, as well as the correct and incorrect predictions for both the \textit{Sup.-only} baseline and our proposed method.

\textbf{Ablation studies.} To test the importance of the NeRF self-supervision and the pseudo-label refinement with the generic SAM masks, we perform an ablation study in which we deactivate each of these two components and report the respective performances. Table \ref{tab:ablationcomp} shows the performance of these two variations of our model compared to the \textit{Sup.-only} baseline and our full model. 

\begin{figure}[h]
    \centering
    \includegraphics[width=0.49\textwidth]{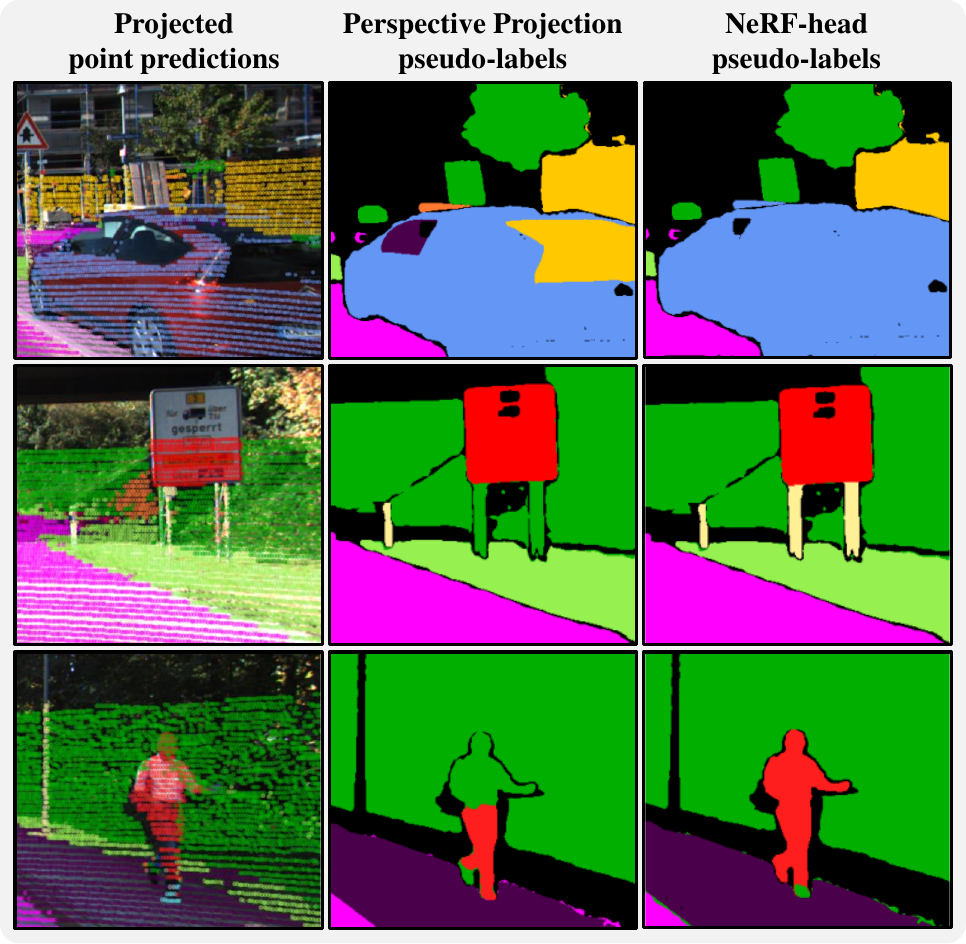}
    \caption{\textbf{Perspective Projection comparison.} Visual examples of the benefit of our NeRF self-supervision compared to the Perspective Projection approach.}
    \label{fig:perspectivequalitative}
\end{figure}  

For the Perspective Projection experiment, we remove the whole NeRF self-supervision, including ray casting, $NeRF$ MLP head, and volumetric rendering. The segment pseudo-labels $\hat{\mathcal{C}}_{s}$ are computed as the $\argmax$ of the semantic predictions from all the projected LiDAR points that fall within each segment's boundaries. Then, we assign 3D pseudo-labels for each prediction from the $vox_{3D}$ head according to the segment pseudo-label $\hat{\mathcal{C}}_{s}$ of its projection. The results show that the performance upgrade with respect to the baseline drops by more than 50\%. Analogously, we show in Fig. \ref{fig:perspectivequalitative} several examples where our NeRF self-supervision outperforms the Perspective Projection when producing 2D pseudo-labels thanks to the ray-wise geometric reasoning which the Perspective Projection approach lacks.

At the No-SAM experiment, we keep the NeRF self-supervision, but instead of producing refined pseudo-labels with the generic SAM masks, we directly supervise each pixel prediction with its $\argmax$. For fairness, we only produce pixel pseudo-labels from the confident pixels. We filter the confident pixel predictions $\hat{y}_p$ by entropy, using the same entropy threshold as the confidence sampler from our full model. The results for this setup show negligible performance upgrade with respect to the \textit{Sup.-only} baseline.

\begin{table}[t]
\addtolength{\tabcolsep}{-1.5pt}%
\caption{Ablation study with the separate contributions of our method in 1\% labeled split of SemanticKITTI \cite{semantickitti}.}
\centering
\begin{tabular}{ cccc }
\toprule
Method & NeRF head & SAM masks  & mIoU (\%) \\
\midrule
\rowcolor{LightCyan}
\textit{Sup.-only} & \xmark & \xmark & 45.4 \\
Perspective Projection & \xmark & \cmark  & 46.4 \\
No-SAM & \cmark & \xmark & 45.8 \\
\midrule
\textbf{Ours} & \cmark & \cmark  & \textbf{47.7} \\
\bottomrule
\end{tabular}
\addtolength{\tabcolsep}{+1.5pt}%
\label{tab:ablationcomp}
\end{table}

\section{CONCLUSION}
This work demonstrates that unlabeled camera images can be used to effectively learn the LiDAR Semantic Segmentation task from unlabeled scans. We show how our NeRF self-supervision effectively closes the domain gap between 2D and 3D data. Unlike methods based on perspective projection, our model is resilient to the parallax effect from the sensor's position shift between captured images and scans. Our results and qualitative examples show that the pixel semantic renderings at the interiors of the objects are confident and reliable. In contrast, non-confident predictions concentrate on the object's boundaries. 
Consequently, our confidence sampling strategy merges the confident renderings with SAM's generic masks to produce reliable pseudo-labels that supervise the non-confident predictions. Our model shows the greatest performance improvements with nuScenes, whose multi-view cameras cover the whole LiDAR's angular range. However, our model produces competitive upgrades even on SemanticKITTI and ScribbleKITTI, where the single frontal camera covers only a small fraction of the LiDAR's angular range. These promising results in all three benchmarks suggest larger performance upgrades when building our multi-modal SSL mechanism on top of other LiDAR-only SSL methods.

\addtolength{\textheight}{-12cm}   


\begin{thebibliography}{99}
\bibitem{dinov1} M. Caron, H. Touvron, I. Misra, H. Jégou, J. Mairal, P. Bojanowski, and A. Joulin, Emerging Properties in Self-Supervised Vision Transformers, ICCV 2021.
\bibitem{dinov2} M. Oquab, T. Darcet, T. Moutakanni, H. Vo, M. Szafraniec, V. Khalidov, P. Fernandez, et al. DINOv2: Learning Robust Visual Features without Supervision, 2023.
\bibitem{sam} A. Kirillov, E. Mintun, N. Ravi, H. Mao, C. Rolland, L. Gustafson, T. Xiao, et al. Segment Anything, ICCV 2023.
\bibitem{samhq} L. Ke, M. Ye, M. Danelljan, Y. Liu, Y. Tai, C. Tang, and F. Yu, Segment Anything in High Quality, NeurIPS 2023.
\bibitem{Roriz2022} R. Roriz, J. Cabral, and T. Gomes, Automotive LiDAR Technology: A Survey, ITSS 2022.
\bibitem{Pendleton2017} S. D. Pendleton, H. Andersen, X. Du, X. Shen, M. Meghjani, Y. H. Eng, D. Rus, and M. H. Ang Perception, Planning, Control, and Coordination for Autonomous Vehicles, MDPI 2017.
\bibitem{SalsaNext} T. Cortinhal, G. Tzelepis, and E. E. Aksoy, SalsaNext: Fast, Uncertainty-aware Semantic Segmentation of LiDAR Point Clouds for autonomous Driving, ISVC 2020.
\bibitem{PolarNet} Y. Zhang, Z. Zhou, P. David, X. Yue, Z. Xi, B. Gong, and H. Foroosh, Polarnet: An improved grid representation for online lidar point clouds semantic segmentation, CVPR 2020.
\bibitem{AMVNet} V. E. Liong, T. N. T. Nguyen, S. Widjaja, D. Sharma, and Z. J. Chong, AMVNet: Assertion-based Multi-View Fusion Network for LiDAR Semantic Segmentation, 2020. 
\bibitem{Tang2020} H. Tang, Z. Liu, S. Zhao, Y. Lin, J. Lin, H. Wang, and S. Han, Searching Efficient 3D Architectures with Sparse Point-Voxel Convolution, ECCV 2020.
\bibitem{semantickitti} J. Behley, M. Garbade, A. Milioto, J. Quenzel, S. Behnke, C. Stachniss, and J. Gall, SemanticKITTI: A Dataset for Semantic Scene Understanding of LiDAR Sequences, ICCV 2019.
\bibitem{scribblekitti} O. Unal, D. Dai, and L. V. Gool, Scribble-Supervised LiDAR Semantic Segmentation, CVPR 2020.
\bibitem{nuscenes} H. Caesar, V. Bankiti, A. H. Lang, S. Vora, V. E. Liong, Q. Xu, A. Krishnan, et al. nuScenes: A multimodal dataset for autonomous driving, CVPR 2020.
\bibitem{waymoopen} P. Sun, H. Kretzschmar, X. Dotiwalla, A. Chouard, V. Patnaik, P. Tsui, J. Guo, et al. Scalability in Perception for Autonomous Driving: Waymo Open Dataset, CVPR 2020.
\bibitem{Mahony2019} N. O. Mahony, S. Campbell, A. Carvalho, L. Krpalkova, D. Riordan, and J. Walsh, Point Cloud Annotation Methods for 3D Deep
Learning, ICST 2019
\bibitem{Ginsburg2008} F. Ginsburg, The Parallax Effect: The Impact of Aboriginal Media on Ethnographic Film. Journal of the Society for Visual Anthropology, 2008.
\bibitem{Schon2009} B. Sch{\"o}n, M. Bertolotto, D. F. Leafer, and S. Morrish, Storage, manipulation, and visualization of LiDAR data, ISPRS 2009.
\bibitem{Hughes2013} J. D. Foley, Computer graphics: principles and practice, 1996.
\bibitem{Genova2021} K. Genova, X. Yin, A. Kundu, C. Pantofaru, F. Cole, A. Sud, B. Brewington, et al. Learning 3D Semantic Segmentation with only 2D Image Supervision, 3DV 2021.
\bibitem{nerf} B. Mildenhall, P. P. Srinivasan, M. Tancik, J. T. Barron, R. Ramamoorthi, and R. Ng, NeRF: Representing Scenes as Neural Radiance Fields for View Synthesis, ACM 2021.
\bibitem{Milioto2019} A. Milioto, I. Vizzo, J. Behley, and C. Stachniss, Rangenet++: Fast and accurate lidar semantic segmentation, IROS 2019.
\bibitem{Xu2021} C. Xu, B. Wu, Z. Wang, W. Zhan, P. Vajda, K. Keutzer, and M. Tomizuka, SqueezeSegV3: Spatially-Adaptive Convolution for Efficient Point-Cloud Segmentation, ECCV 2020.
\bibitem{Zhao2021} Y. Zhao, L. Bai, and X. Huang, FIDNet: LiDAR Point Cloud Semantic Segmentation with Fully Interpolation Decoding, IROS 2021.
\bibitem{Chen2017} L. C. Chen, G. Papandreou, F. Schroff, and H. Adam, Rethinking Atrous Convolution for Semantic Image Segmentation, CVPR 2017.
\bibitem{PointNet} C. R. Qi, H. Su, K. Mo, and L. J. Guibas, PointNet: Deep Learning on Point Sets for 3D Classification and Segmentation, CVPR 2017.
\bibitem{Zhang2020} Y. Zhang, Z. Zhou, P. David, X. Yue, Z. Xi, B. Gong, and H. Foroosh, PolarNet: An Improved Grid Representation for Online LiDAR Point Clouds Semantic Segmentation, CVPR 2020.
\bibitem{RPVNet} J. Xu, R. Zhang, J. Dou, Y. Zhu, J. Sun, and S. Pu, RPVNet: A Deep and Efficient Range-Point-Voxel Fusion Network for LiDAR Point Cloud Segmentation, ICCV 2021.
\bibitem{Zhao2020} H. Tang, Z. Liu, S. Zhao, Y. Lin, J. Lin, H. Wang, and S. Han, Searching Efficient 3D Architectures with Sparse Point-Voxel Convolution, ECCV 2020.
\bibitem{Cylinder3D2020} H Zhou, X Zhu, X Song, Y Ma, Z Wang, H. Li, and D. Lin, Cylinder3D: An Effective 3D Framework for Driving-scene LiDAR Semantic Segmentation, 2020.
\bibitem{Cylinder3D2021} X. Zhu, H. Zhou, T. Wang, F. Hong, W. Li, Y. Ma, H. Li, et al. Cylindrical and Asymmetrical 3D Convolution Networks for LiDAR-based Perception, TPAMI 2021.
\bibitem{Gao2020} B. Gao, Y. Pan, C. Li, S. Geng, and H. Zhao, Are We Hungry for 3D LiDAR Data for Semantic Segmentation? A Survey and Experimental Study, T-ITS 2021.
\bibitem{SQN} Q. Hu, B. Yang, G. Fang, Y. Guo, A. Leonardis, N. Trigoni, and A. Markham, SQN: Weakly-Supervised Semantic Segmentation of Large-Scale 3D Point Clouds, ECCV 2022.
\bibitem{Zhang2021} Y. Zhang, Y. Qu, Y. Xie, Z. Li, S. Zheng, and C. Li, Perturbed Self-Distillation: Weakly Supervised Large-Scale Point Cloud Semantic Segmentation, ICCV 2021.
\bibitem{Box2Seg} Y. Liu, Q. Hu, Y. Lei, K. Xu, J. Li, and Y. Guo, Box2Seg: Learning Semantics of 3D Point Clouds with Box-Level Supervision, 2022.
\bibitem{Chen2021} X. Chen, Y. Yuan, G. Zeng, and J. Wang, Semi-Supervised Semantic Segmentation with Cross Pseudo Supervision, CVPR 2021.
\bibitem{Ouali2020} Y. Ouali, C. Hudelot, and M. Tami, Semi-Supervised Semantic Segmentation with Cross-Consistency Training, CVPR 2020.
\bibitem{MeanTeacher} A. Tarvainen and H. Valpola, Mean teachers are better role models: Weight-averaged consistency targets improve semi-supervised deep learning results, NeurIPS 2017.
\bibitem{Ke2020} Z. Ke, D. Qiu, K. Li, Q. Yan, and R. W. H. Lau, Guided Collaborative Training for Pixel-wise Semi-Supervised Learning, ECCV 2020.
\bibitem{Lee2013} D. H. Lee, Pseudo-Label: D.H. Lee, The Simple and Efficient Semi-Supervised Learning Method for Deep Neural Networks, ICML 2013.
\bibitem{Hu2021} H. Hu, F. Wei, H. Hu, Q. Ye, J. Cui, and L. Wang, Semi-Supervised Semantic Segmentation via Adaptive Equalization Learning, NeurIPS 2021.
\bibitem{Yang2022} L. Yang, W. Zhuo, L. Qi, Y. Shi, and Y. Gao, ST++: Make Self-training Work Better for Semi-supervised Semantic Segmentation, CVPR 2022
\bibitem{Yuan2021} J. Yuan, Y. Liu, C. Shen, Z. Wang, and H. Li. A Simple Baseline for Semi-supervised Semantic Segmentation with Strong Data Augmentation, ICCV 2021.
\bibitem{LaserMix} L. Kong, J. Ren, L. Pan, and Z. Liu, LaserMix for Semi-Supervised LiDAR Semantic Segmentation, CVPR 2023.
\bibitem{Bang2021} D. Bang, J. Lee, and H. Shim, Distilling from professors: Enhancing the knowledge distillation of teachers, Information Sciences 2021.
\bibitem{Zhou2023} C. Zhou, Q. Li, C. Li, J. Yu, Y. Liu, G. Wang, K. Zhang, et al. A Comprehensive Survey on Pretrained Foundation Models: A History from BERT to ChatGPT, 2023.
\bibitem{Huang2022} K. Huang, B. Shi, X. Li, X. Li, S. Huang, and Y. Li, Multi-modal Sensor Fusion for Auto Driving Perception: A Survey, 2022.
\bibitem{FCN} J. Long, E. Shelhamer, and T. Darrell, Fully Convolutional Networks for Semantic Segmentation, CVPR 2015.
\bibitem{Lovasz} M. Berman, A. R. Triki, and M. B. Blaschko, The Lovasz-Softmax loss: A tractable surrogate for the optimization of the intersection-over-union measure in neural networks, CVPR 2018.
\bibitem{CBST} Y. Zou, Z. Yu, B. V. K. Kumar, and J. Wang, Unsupervised Domain Adaptation for Semantic Segmentation via Class-Balanced Self-Training, ECCV 2018.
\bibitem{CutMix} G. French, S. Laine, T. Aila, M. Mackiewicz, and G. Finlayson, Semi-supervised semantic segmentation needs strong, varied perturbations, ICCV 2019.
\bibitem{CPS} X. Chen, Y. Yuan, G. Zeng, and J. Wang, Semi-Supervised Semantic Segmentation with Cross Pseudo Supervision, CVPR 2021.
\bibitem{Xie2016} J. Xie, M. Kiefel, M. T. Sun, and A. Geiger, Semantic Instance Annotation of Street Scenes by 3D to 2D Label Transfer, CVPR 2016.
\bibitem{Kong2023} L. Kong, Y. Liu, R. Chen, Y. Ma, X. Zhu, Y. Li, Y. Hou, Y. Qiao, and Z. Liu, Rethinking Range View Representation for LiDAR Segmentation, ICCV 2023.
\end{thebibliography}
\end{document}